\documentclass{article}
\usepackage{acra}
\usepackage{graphicx}
\usepackage{amsmath,amssymb}
\usepackage{booktabs}
\usepackage{array}
\usepackage{tikz}
\usetikzlibrary{arrows.meta,positioning}
\usepackage{subcaption}
\tikzset{
dim/.style={{Stealth[length=1.5mm]}-{Stealth[length=1.5mm]}, thin}, ext/.style={densely dashed, gray, thin}, move/.style={-{Stealth[length=2.6mm]}, line width=1.1pt}, force/.style={-{Stealth[length=2.4mm]}, line width=1.4pt, black}, target/.style={draw, thick, fill=black!5}, massb/.style={draw, thick, fill=blue!10}, }
\newcommand{\annotfont}{\small}

\title{Flattening Fitts' Law: Difficulty-Indexed Haptic Assistance for Reaching Tasks}

\author{Andrew Razjigaev$^{*\dagger}$, Nguyen B.N. Le$^{*}$, Jonathan Eden, \\ \textbf{Vincent Crocher, Wafa Johal, Denny Oetomo }\\
Department of Mechanical Engineering, School of Computing and Information Systems \\ The University of Melbourne, Melbourne, Australia}

\begin{document}
\maketitle
{\renewcommand{\thefootnote}{\fnsymbol{footnote}}\footnotetext[1]{These authors contributed equally to this work.}
\footnotetext[2]{Corresponding author: \\  \indent
\texttt{ andrew.razjigaev@unimelb.edu.au}}
}

\begin{abstract}
Rehabilitation robots physically assist patients through movements, where the level of assistance must balance two risks: too little leaves the patient \textcolor{black}{struggling} to bear the full task difficulty, while too much invites motor slacking that erodes their participation. Assistance is commonly adapted from tracking error. However, tracking error reports that a movement has already strayed rather than how demanding the task is before it begins. We propose indexing assistance to task difficulty, placing the human and the assisting robot on a common Fitts' law-based scale and modelling their shared movement time as a linear blend. This gives a closed-form arbitration weight $\alpha$ that looks to hold movement time at a target independent of difficulty. We evaluate this approach in a planar reaching study with 20 participants. \textcolor{black}{Robotic assistance} \textcolor{black}{allows a modulation of} the Fitts movement time with respect to difficulty. However, \textcolor{black}{robotic-assisted} reaching movements settled 16\% faster than the target, even after accounting for practice. Moreover, trials were often faster than the robot alone, indicating an interaction between the human and robot that the blend omits. Participants rated the robot as fluent and successfully understood increases in autonomy. This difficulty-indexed assistance therefore offers an interpretable starting point for future control, provided the user's own contribution is folded back into the arbitration.
\end{abstract}

\section{Introduction}

Robot-assisted therapy is a promising route to restoring movement after neurological injury: a robot can support a patient's limb through the intensive, repetitive practice that recovery demands \cite{mehrholz_electromechanical_2018}. The effectiveness of rehabilitation robots rests partly on how well the challenge of the exercise is matched to the patient's current ability. Here, flow theory holds that a person is most engaged, and learns most readily, when the challenge of a task is balanced against their skill: too much challenge brings anxiety and failure, too little brings boredom and disengagement \cite{nakamura2002flow}. The same balance underlies motor learning, which is most effective at an intermediate, individualised difficulty \cite{guadagnoli2004}. Effective assistance should therefore keep the patient within this flow channel, attenuating challenges that they cannot yet meet and withdrawing support as skill grows, so that practice stays demanding enough to drive progression.

Delivering this balance makes assistance an adaptation problem. Assist-as-needed controllers lower support with improvement \cite{marchalcrespo2009}, and systems that jointly tune task difficulty and haptic assistance can hold the challenge-skill ratio near its optimum \cite{rodriguezguerrero2017,zimmerli2012}. The adaptation must not be too little or too much: too little leaves the patient to bear the full difficulty, whereas too much invites a reliance on the robot (\textit{motor slacking}) which removes the active participation on which learning depends \cite{reinkensmeyer2009slacking,milot2010}. Rehabilitation here is one instance of the shared-control problem, common to teleoperation and skill training, of arbitrating authority between a human and an assisting controller \cite{dragan2013,losey2018}.

Adapting assistance well requires two ingredients: a model of the user's capability and a model of task difficulty \cite{guadagnoli2004}. The second is often left implicit. Assistance is commonly set by heuristic, or adapted from tracking error, performance, or psychophysiological signals \cite{marchalcrespo2009,rodriguezguerrero2017}. However, these are reactive, reporting that a movement has already deviated rather than how demanding the task is in advance. Tracking error, in particular, is an ambiguous indicator. Here, a large error may signal a hard task, but equally a distraction or an alternative but valid path. A difficulty model such as Fitts' law instead gives an explicit, a-priori model of task demand: the movement time of a reach rises linearly with an index of difficulty that combines the distance to the target and its width \cite{fitts1954,mackenzie1992}, fixed by the task geometry and therefore is known before the movement begins. This relationship is reproducible enough to underpin the ISO~9241-9 standard for pointing evaluation \cite{soukoreff2004} and has recently been used to benchmark how difficulty and robot autonomy affect assisted human-robot reaching \cite{pan2025fitts}. We build on this to derive a law for assistance and investigate its performance.

We therefore use a task difficulty model to set assistance. Placing the human and the robot on a common Fitts' law difficulty scale, we ask: \emph{given a model of how difficult a reach is, how can a robot set its assistance to reduce the difficulty that the human experiences towards a target value?} We look to answer this with a single arbitration weight, computed from the task-difficulty model, that attenuates the reach difficulty by a prescribed amount, giving a direct and interpretable means of moving a user within the flow channel. This treats the human as a passive term; yet under assistance an operator may disengage and slack, or over-engage. We therefore ask a second, empirical question: \emph{does the assisted human-robot interaction follow the linear blend, or does the operator's engagement make the realised movement time depart from it?} Our contributions are:


\begin{enumerate}
\item \textbf{Indexing Assistance by Task Difficulty}: we set the arbitration weight $\alpha$ from the geometric task difficulty, known before the movement, rather than from tracking error. This is achieved by placing the human and robot on a common Fitts' law scale and modelling the assisted movement time as a linear blend of the two. The blend treats the human and robot as independent, making it a falsifiable passive-human baseline rather than an assumed result.
\item \textbf{An Experimental Evaluation}: we characterise the human and robot Fitts' law relationships on a planar manipulandum and measure how the assisted movement time departs from the baseline, anticipating an active departure due to the participants' response to the interaction \cite{pan2025fitts}. We also explore how participants perceive its fluency and level of autonomy.
\end{enumerate}
By setting assistance via task difficulty rather than tracking error, this offers a principled way to vary a task's challenge toward a user's ability, which is of direct use in rehabilitation and motor-skill training.

\section{Related Work}

Fitts' law provides a quantitative measure of reaching difficulty. It models aimed movement as an information-transmission process in which movement time $MT$ is linear with respect to an index of difficulty $ID$ \cite{fitts1954,mackenzie1992},
\begin{equation}
MT=a+b\,ID,
\end{equation}
with fitted constants $a$ and $b$. 
As this relationship has been shown to hold across devices and users, the slope reciprocal $1/b$ serves as a throughput, and the law underpins the ISO~9241-9 standard for pointing evaluation \cite{soukoreff2004}. The law also admits a control-theoretic reading: a first-order feedback loop nulling its error to a given tolerance takes a time that is logarithmic in distance-to-tolerance \cite{crossman1983}, refined by optimal sub-movement models \cite{meyer1988}.

Shared control can arbitrate authority between the human and the autonomous controller in several ways, from switching or trading control between them, to constraining motion with virtual fixtures, to continuously blending their two control signals \cite{abbink2018,losey2018}. The last is particularly attractive: the convex combination of control inputs
\begin{equation}
u=\alpha\,u_r+(1-\alpha)\,u_h,\qquad \alpha\in[0,1], \label{eq:policyblend}
\end{equation}
is simple, continuous and interpretable, a single weight $\alpha$ spanning pure-human $u_h$ ($\alpha=0$) to pure-robot $u_r$ ($\alpha=1$) control input, so that arbitration reduces to choosing that weight \cite{dragan2013}. How to set $\alpha$, whether from predicted intent, confidence, or task state, remains an unresolved question in physical human-robot interaction \cite{losey2018,abbink2018}. Closest to our aim, Pan et al.\ used Fitts' law to benchmark assisted reaching, finding that movement time decreased with greater robot autonomy and, notably, that operators moved at higher velocities and that the unassisted Fitts' law trend was altered under assistance \cite{pan2025fitts}. However, that work fixed the assistance and measured its effect, rather than prescribing it. Arbitration is thus rarely tied to a validated model of task difficulty; we instead investigate a method for deriving the arbitration weight $\alpha$ that achieves a target movement time.

In rehabilitation robotics, assistance is adapted to the patient in several ways. Assist-as-needed controllers reduce support as performance improves, but are usually keyed to tracking error or a forgetting factor \cite{marchalcrespo2009}; over-assistance then invites slacking, reducing the active participation that drives motor learning \cite{reinkensmeyer2009slacking,milot2010}. Others adapt the task itself, reshaping target size, distance or allowed time to the user's capability \cite{zimmerli2012}, or tune difficulty and assistance jointly from performance and psychophysiological signals to hold the challenge-skill ratio near its optimum \cite{rodriguezguerrero2017}. Difficulty is the lever these methods share, since learning is greatest at an intermediate, individualised difficulty \cite{guadagnoli2004}; yet that difficulty is usually inferred from the user's performance rather than from the task's features. Adapting assistance well therefore calls for both a model of the user's capability and a model of task difficulty. We contribute the latter: holding the task fixed and adapting the assistance force, we shift the realised difficulty through the convex combination of the human and robot movement times, so that assistance is indexed by difficulty rather than by error, and departures from the model, whether slacking or feedforward, ideally become measurable rather than an afterthought.

\section{Difficulty-indexed Assistance Modelling}
\label{sec:model}

\begin{figure}[t]
\centering
\begin{subfigure}[b]{\columnwidth}
\centering
\begin{tikzpicture}[x=0.82cm, y=0.82cm, font=\small]
\begin{scope}
  \def\A{7.5}\def\Wh{1.0}
  \node[target, minimum width={2*\Wh cm/1.22}, minimum height=1cm] at (\A,0) {};
  \node at (\A,0) {Target};
  \draw[ext] (\A,-0.5) -- (\A,-0.2);
  \fill[red!70!black] (0,0) circle (2.2pt);
  \node[anchor=south] at (0,0.28) {Start};
  \draw[move] (0.15,0) -- (\A-\Wh,0);
  \draw[ext] (\A-\Wh,0.5) -- (\A-\Wh,1.5);
  \draw[ext] (\A+\Wh,0.5) -- (\A+\Wh,1.5);
  \draw[dim] (\A-\Wh,1.35) -- (\A+\Wh,1.35) node[midway,above]{$W$};
  \draw[ext] (0,0) -- (0,-1.5);
  \draw[ext] (\A,-0.5) -- (\A,-1.5);
  \draw[dim] (0,-1.35) -- (\A,-1.35) node[midway,below]{$A$};
\end{scope}
\end{tikzpicture}
\caption{Fitts' law reaching paradigm}
\label{fig:fitslaw}
\end{subfigure}
\\
\begin{subfigure}[b]{\columnwidth}
\centering
\begin{tikzpicture}[x=0.82cm, y=0.82cm, font=\small]
\begin{scope}[shift={(0,-5.2)}]
  \def\A{7.5}\def\Wh{1.0}\def\xpos{3.4}
  \node[target, minimum width={2*\Wh cm/1.22}, minimum height=1cm] at (\A,0) {};
  \node at (\A,0) {Target};
  \draw[ext] (\A,-0.95) -- (\A,-0.2);
  \draw[ext] (\A-\Wh,0.5) -- (\A-\Wh,1.5);
  \draw[ext] (\A+\Wh,0.5) -- (\A+\Wh,1.5);
  \draw[dim] (\A-\Wh,1.35) -- (\A+\Wh,1.35) node[midway,above]{$W$};
  \draw[ext] (\A-\Wh,-0.5) -- (\A-\Wh,-0.99);
  \draw[dim] (\A-\Wh,-0.78) -- (\A,-0.78) node[midway,below,yshift=1pt]{\scriptsize $W/2$};
  \node[massb, minimum size=0.62cm] at (0,0) {$m$};
  \draw[force] (0.42,0) -- (1.7,0) node[midway,above]{$u$};
  \node[draw, thick, fill=blue!4, minimum size=0.6cm] at (\xpos,0) {};
  \draw[ext] (\xpos,0.4) -- (\xpos,1.0);
  \draw[dim] (\xpos,0.9) -- (\A,0.9) node[midway,above]{$e=A-x$};
  \draw[-{Stealth[length=2mm]}] (-0.4,-1.5) -- (8.4,-1.5) node[right]{$x$};
  \draw (0,-1.4) -- (0,-1.6) node[below]{$0$};
  \draw[ext] (\xpos,-0.3) -- (\xpos,-1.5);
  \draw[dim] (0,-1.0) -- (\xpos,-1.0) node[midway,below,yshift=1pt]{$x$};
\end{scope}
\end{tikzpicture}
\caption{Robot model: mass $m$ driven by force $u$}
\label{fig:robotmodel}
\end{subfigure}
\caption{Target reaching with width $W$ and amplitude $A$ (measured to the target centre). (a)~The human reaches from the start; movement time follows Fitts' law~\eqref{eq:mth}. (b)~The robot drives the handle (mass $m$) to the centre by an input $u=K_p e+K_d\dot e$ with error $e=A-x$.}
\label{fig:setup}
\end{figure}
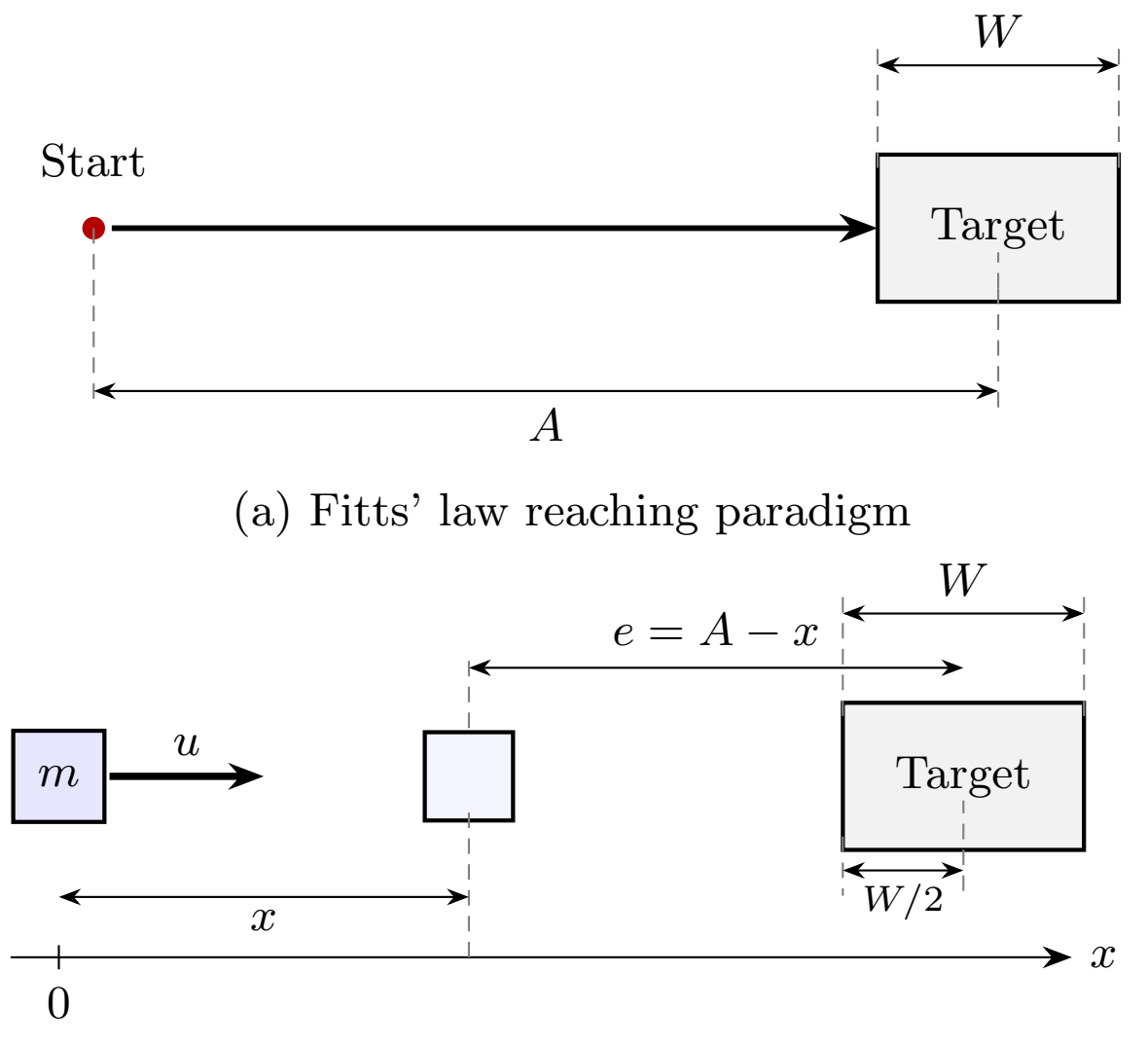

\subsection{Model of Human and Robot Difficulty}
For a target of width $W$ at amplitude $A$ (Fig.~\ref{fig:fitslaw}), the Shannon index of difficulty is
\begin{equation}
ID=\log_2\!\left(\frac{A}{W}+1\right),
\label{eq:shannon}
\end{equation}
and human movement time follows Fitts' law
\begin{equation}
MT_h=a_h+b_h\,ID, \label{eq:mth}
\end{equation}
with participant-specific intercept $a_h$ and slope $b_h$.

We now model the robot driving the handle (Fig.~\ref{fig:robotmodel}) as a mass $m$ under a proportional-derivative (PD) regulator such that:

\begin{equation}
    u_r=K_p\,e+K_d\,\dot e, \label{eq:pd}
\end{equation}
where $e$ is the error $e(t) = A - x(t)$ with respect to the reference position $A$. 
Friction compensation is always active on the device so that the handle approximates the ideal mass $m$, and is omitted from the model.

A stable closed loop controller drives the error to zero exponentially, which can be modelled as $e(t)\approx A\,e^{-t/\tau}$, with time constant $\tau$ set by $K_p,K_d$ \cite{astrom2021}. The reach is complete when the end-effector first enters the target,  after traversing the minimum distance $A-W/2$. Setting $e(t)=W/2$ gives the acquisition time $t=\tau\ln(2A/W)$, the classical first-order feedback account of Fitts' law \cite{crossman1983}: acquisition time is logarithmic in $2A/W$, so the robot movement time is, in principle, Fitts-linear:
\begin{equation}
MT_r = a_r+b_r\,\log_2\!\frac{2A}{W},\qquad b_r\approx\tau\ln 2. \label{eq:mtr}
\end{equation}
With a slope far smaller than the human's, $b_r\ll b_h$, the robot is comparatively insensitive to difficulty, always reaching the target faster than the human ($MT_r<MT_h$). Eq.~\eqref{eq:mtr} is a first-order approximation, so rather than predict $a_r$ and $b_r$ from the device parameters we subsequently estimate them empirically using the robot's performance in the reaching task. 

\subsection{A Linear-blend Baseline}
\begin{figure}[t]
\centering
\begin{tikzpicture}[x=0.9cm, y=0.82cm, font=\small,
  ax/.style={-{Stealth[length=2.8mm]}, line width=1.2pt},
  ml/.style={line width=1.1pt}]
  \def\xa{0.15}\def\xb{4.7}
  \fill[black!7] (\xa,{0.22+0.13*\xa}) -- (\xb,{0.22+0.13*\xb}) --
                 (\xb,{0.45+0.52*\xb}) -- (\xa,{0.45+0.52*\xa}) -- cycle;
  \draw[ax] (-0.1,0) -- (6.6,0) node[below]{$ID$};
  \draw[ax] (0,-0.1) -- (0,3.7) node[left]{$MT$};
  \draw[ml,red!75!black]  (\xa,{0.45+0.52*\xa}) -- (\xb,{0.45+0.52*\xb})
        node[right,black]{$MT_h=a_h+b_h\,ID$};
  \draw[ml,blue!70!black] (\xa,{0.22+0.13*\xa}) -- (\xb,{0.22+0.13*\xb})
        node[right,black]{$MT_r=a_r+b_r\,ID$};
  \node[red!75!black]  at (1.7,{0.45+0.52*1.7+0.42}) {$\alpha=0$};
  \node[blue!70!black] at (3.7,{0.22+0.13*3.7-0.36}) {$\alpha=1$};
  \def\C{0.8}\def\idL{0.673}\def\idR{4.46}
  \draw[gray, densely dotted] (\idL,\C) -- (\idL,0);
  \draw[gray, densely dotted] (\idR,\C) -- (\idR,0);
  \draw (\idL,0.08) -- (\idL,-0.08) node[below, font=\footnotesize]{$ID_{\min}$};
  \draw (\idR,0.08) -- (\idR,-0.08) node[below, font=\footnotesize]{$ID_{\max}$};
  \draw[ml, teal, densely dashed] (\idL,\C) -- (\idR,\C);
  \fill[teal] (\idL,\C) circle (1.6pt);
  \fill[teal] (\idR,\C) circle (1.6pt);
  \node[teal] at (3.55,{\C+0.32}) {$MT_o^\star$};
\end{tikzpicture}
\caption{The blending model. The human ($\alpha=0$) and robot ($\alpha=1$) are each Fitts-linear, and blended via Eq.~\eqref{eq:blend} to yield a value within the shaded envelope. Letting $\alpha$ vary with difficulty can achieve any value within the envelope, such as the constant set point $MT_o^\star$ (teal), which holds the MT difficulty-invariant by raising $\alpha(ID)$ from $0$ on the human line to $1$ on the robot line.}
\label{fig:blend}
\end{figure}
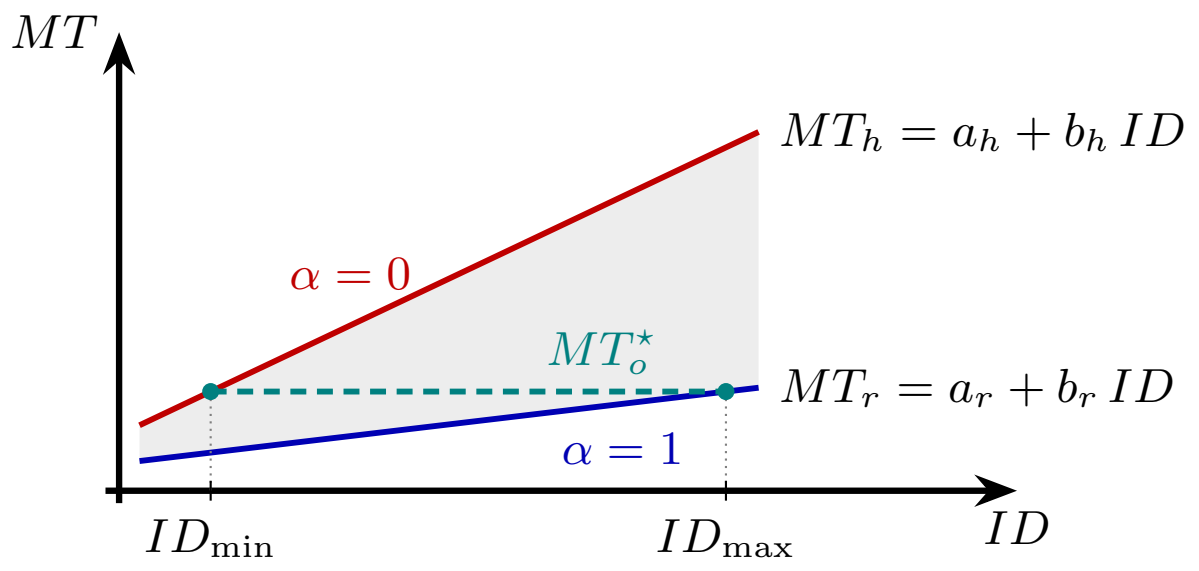

Following the policy-blending formalism \cite{dragan2013}, we model the assisted MT as a blend of the two MTs with an arbitration weight $\alpha\in[0,1]$:
\begin{equation}
MT_o=\alpha\,MT_r+(1-\alpha)\,MT_h. \label{eq:blend}
\end{equation}
This assumes the human and robot act independently: each contributes the MT it achieves alone, with no coupling between them, such that the human is a passive term in the blend. Over the stimulus range, $\log_2(2A/W)$ and the Shannon $ID$ are near-affinely related \cite{mackenzie1992}, so we take $MT_r\approx a_r+b_r\,ID$ and place both lines on one $ID$ axis (Fig.~\ref{fig:blend}). As a convex combination,~\eqref{eq:blend} bounds the assisted time between the two solo lines, $MT_r\le MT_o\le MT_h$, with $\alpha=0$ giving the human and $\alpha=1$ the robot. The robot time is therefore the performance ceiling of the model: a measured $MT_o<MT_r$ would falsify the linear blend, indicating a human-robot coupling that the convex combination cannot represent.

\subsection{Difficulty-varying Assistance}
Given a desired target $MT_o^\star$ at each condition, inverting~\eqref{eq:blend} gives the pointwise weight
\begin{equation}
\alpha(ID)=\frac{MT_o^\star-MT_h(ID)}{MT_r(ID)-MT_h(ID)}, \label{eq:alpha_id}
\end{equation}
Since the robot is faster than the human over the tested range ($MT_r<MT_h$), any target within $MT_r\le MT_o\le MT_h$ gives $\alpha\in[0,1]$.
The schedule~\eqref{eq:alpha_id} increases with difficulty, adding enough robot authority at each difficulty to cancel the human's difficulty sensitivity. The realised $MT_o$ is flat only if the human and robot act independently as~\eqref{eq:blend} assumes, so a residual difficulty dependence or offset can reveal a human-robot coupling that the blend omits.

\subsection{Blend Implementation}
The policy blend~\eqref{eq:policyblend} cannot be directly applied since the handle simply sums the force the robot commands, $u_c$, and the force the human applies, $u_h$, giving a net force $u=u_c+u_h$. The device is transparent by default (with static-friction compensation always applied), so at $u_c=0$ the operator feels only the ideal mass. Requiring the net force to equal the blended input~\eqref{eq:policyblend} and solving for the commanded force gives
\begin{equation}
u_c=\alpha\,(u_r-u_h),\qquad u_r=K_p\,e+K_d\,\dot e, \label{eq:force}
\end{equation}
with a hard clamp $\lVert u_c\rVert\le f_{\max}$ for safety. The controller therefore adds a fraction $\alpha$ of the robot input while subtracting that fraction of the measured human force, so that the net force on the handle is the policy blend \eqref{eq:policyblend}.
This realises the input blend of \cite{dragan2013}, at the cost of requiring the interaction force $u_h$ to be measured; stable authority-sharing on a coupled device whose command depends on that force is itself an active question \cite{guo2025passivity}.

\paragraph{A prediction, not an identity.}
Applying~\eqref{eq:force} realises the instantaneous force blend exactly; equating the resulting movement time to the blend~\eqref{eq:blend} is a further step that assumes the operator acts independently of the robot. The human is not a fixed policy, and under assistance may slack or co-contract \cite{reinkensmeyer2009slacking}, while the coupled dynamics are nonlinear. 
We therefore treat the linear blend as a falsifiable passive-human baseline rather than a relationship we expect to hold exactly; the evidence that assisted operators actively speed up \cite{pan2025fitts} leads us to anticipate a feedforward departure ($MT_o<MT_o^\star$), and the direction and size of any departure are the result of interest.

\section{Experiment}
The experiment looks to characterise the human and robot Fitts' law relationships and apply the difficulty-indexed assistance of Section~\ref{sec:model}. It addresses three research questions. \emph{RQ1: Does the difficulty-varying weight $\alpha(ID)$ of~\eqref{eq:alpha_id} make assisted movement time equal to each participant's target $MT_o^\star$ at every $ID$?} The linear blend~\eqref{eq:blend} predicts this; we hypothesise that assistance removes the increase of movement time with $ID$ seen without assistance (H1A), but that $MT_o$ departs from $MT_o^\star$, falling below it as the operator adds feedforward input \cite{pan2025fitts} (H1B). \emph{RQ2: Is difficulty-indexed assistance perceived as fluent, with the robot contributing to the task?} We hypothesise that participants rate both the fluency of the interaction and the robot's contribution above neutral (H2). \emph{RQ3: Do participants perceive the level of autonomy that the robot applies?} We hypothesise that perceived autonomy matches the applied weight $\alpha$ (H3).

\subsection{Participants}
20 healthy adults (13 male, 7 female; aged $23.6 \pm 2.7$ years, mean $\pm$ SD) took part. 19 self-reported as right-handed; one reported a left-hand preference but, like all other participants, performed the task with the right hand, for which the setup was configured. The study was approved by the Human Research Ethics Committee at the University of Melbourne (Ethics ID 15695). All participants gave written informed consent. 


\subsection{Setup}
\begin{figure}[t]
\centering
\begin{subfigure}[b]{\columnwidth}
\centering
\begin{tikzpicture}
\node[anchor=south west,inner sep=0] (ui) at (0,0)
     {\includegraphics[width=0.92\columnwidth,trim=215 220 370 170,clip]{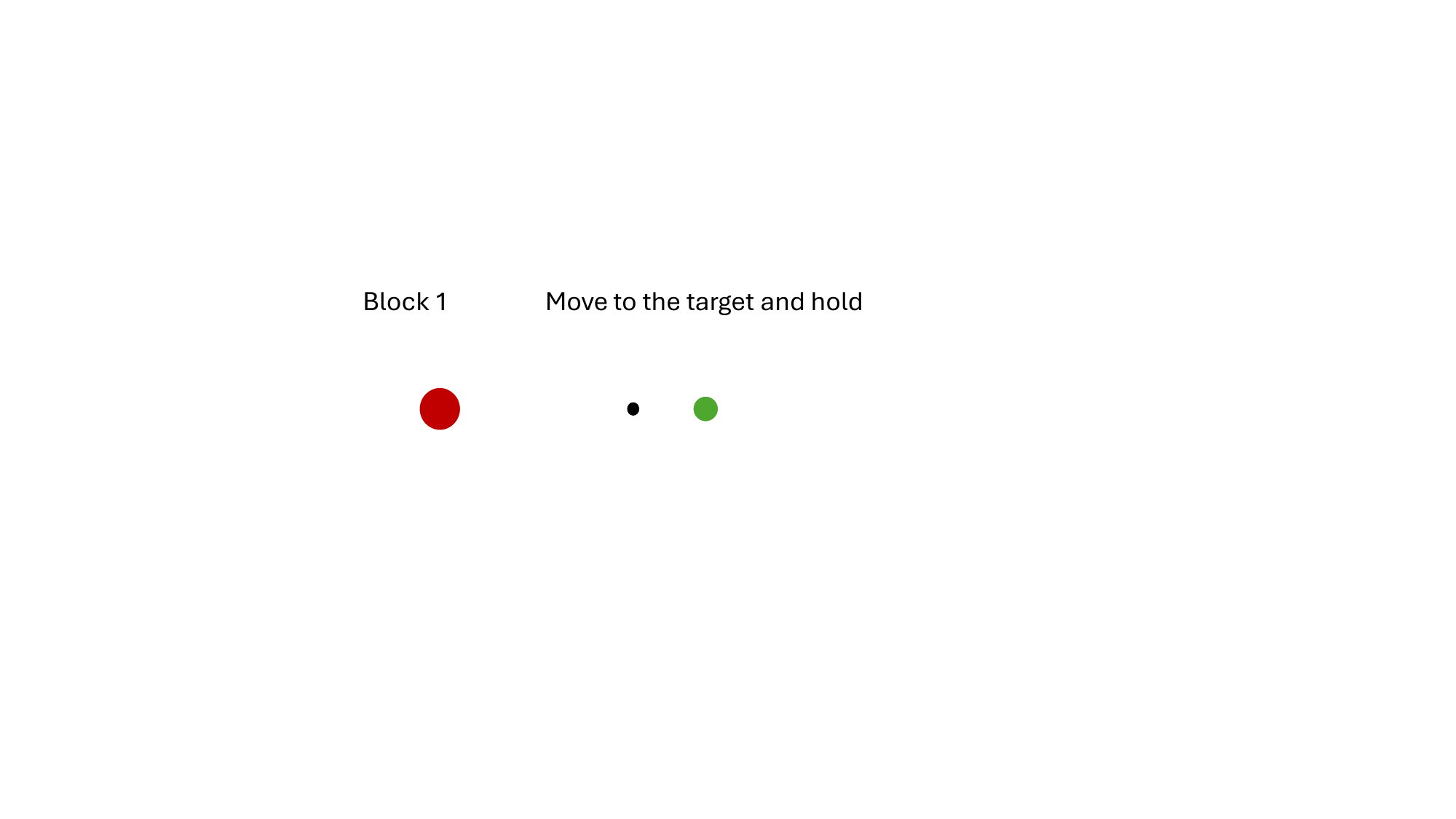}};
\draw[black!35, line width=0.6pt] (ui.south west) rectangle (ui.north east);
\begin{scope}[x={(ui.south east)},y={(ui.north west)}]
  \draw[-{Stealth[length=2mm]}, densely dashed, thick] (0.50,0.338) -- (0.245,0.338);
  \node[font=\annotfont,anchor=north] at (0.198,0.20) {Target};
  \node[font=\annotfont,anchor=north] at (0.532,0.20) {Cursor};
  \node[font=\annotfont,anchor=north] at (0.627+0.05,0.20) {Start};
\end{scope}
\end{tikzpicture}
\caption{Participant's display (Unity)}
\label{fig:unity}
\end{subfigure}
\\[3pt]
\begin{subfigure}[b]{\columnwidth}
\centering
\begin{tikzpicture}
\node[anchor=south west,inner sep=0] (img) at (0,0)
     {\includegraphics[width=\columnwidth]{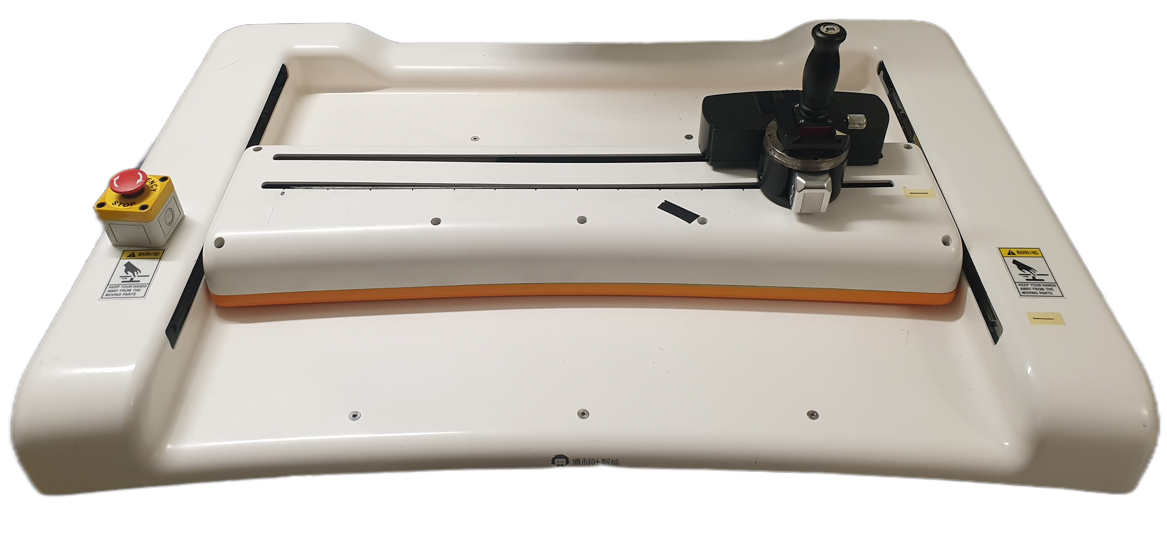}};
\begin{scope}[x={(img.south east)},y={(img.north west)}]
  \def\ox{0.70}\def\oy{0.665}   
  \def\cmS{0.0135}               
  \def\dir{-1}                   
  \draw[-{Stealth[length=3mm]}, cyan!55!black, line width=1.3pt]
       (\ox,\oy) -- ({\ox+\dir*31.5*\cmS},\oy)
       node[above=2.5pt,font=\annotfont,text=black]{$x$};
  \foreach \a/\lab in {0/0,10/10,20/20,30/30\,cm}{
     \pgfmathsetmacro\tx{\ox+\dir*\a*\cmS}
     \draw[cyan!55!black,line width=1pt] (\tx,{\oy+0.022}) -- (\tx,{\oy-0.022});
     \node[font=\annotfont,text=black,anchor=north] at (\tx,{\oy-0.05}) {\lab};
  }
  \foreach \a in {8,12,15,15.1875,18,20.25,27,30.375}{
     \pgfmathsetmacro\tx{\ox+\dir*\a*\cmS}
     \fill[red!75!black] (\tx,\oy) circle (1.3pt);
  }
  \draw[-{Stealth[length=3mm]}, orange!80!black, line width=1.3pt]
       (\ox,\oy) -- (\ox,{\oy+0.20}) node[right=1pt,font=\annotfont,text=black]{$y$};
  \node[font=\annotfont,text=black,anchor=east] (hlab) at (0.63,1.03) {Handle (end-effector)};
  \draw[-{Stealth[length=2mm]}, thick] (hlab.east) -- (0.72,0.86);
\end{scope}
\end{tikzpicture}
\caption{ArmMotus~M2 device}
\label{fig:m2device}
\end{subfigure}
\caption{Experimental setup. (a) The display: the participant moves the cursor (handle position) from the start circle to the target and holds it there. (b) The ArmMotus~M2 planar robot. The handle moves in the horizontal plane over a $0.625\times0.440$~m$^2$ workspace. Reaching is restricted to one $x$-axis direction; dots mark the target amplitudes of Table~\ref{tab:conditions}, spanning $8$ to $30.4$~cm from the $0$~cm start point.}
\label{fig:m2setup}
\end{figure}
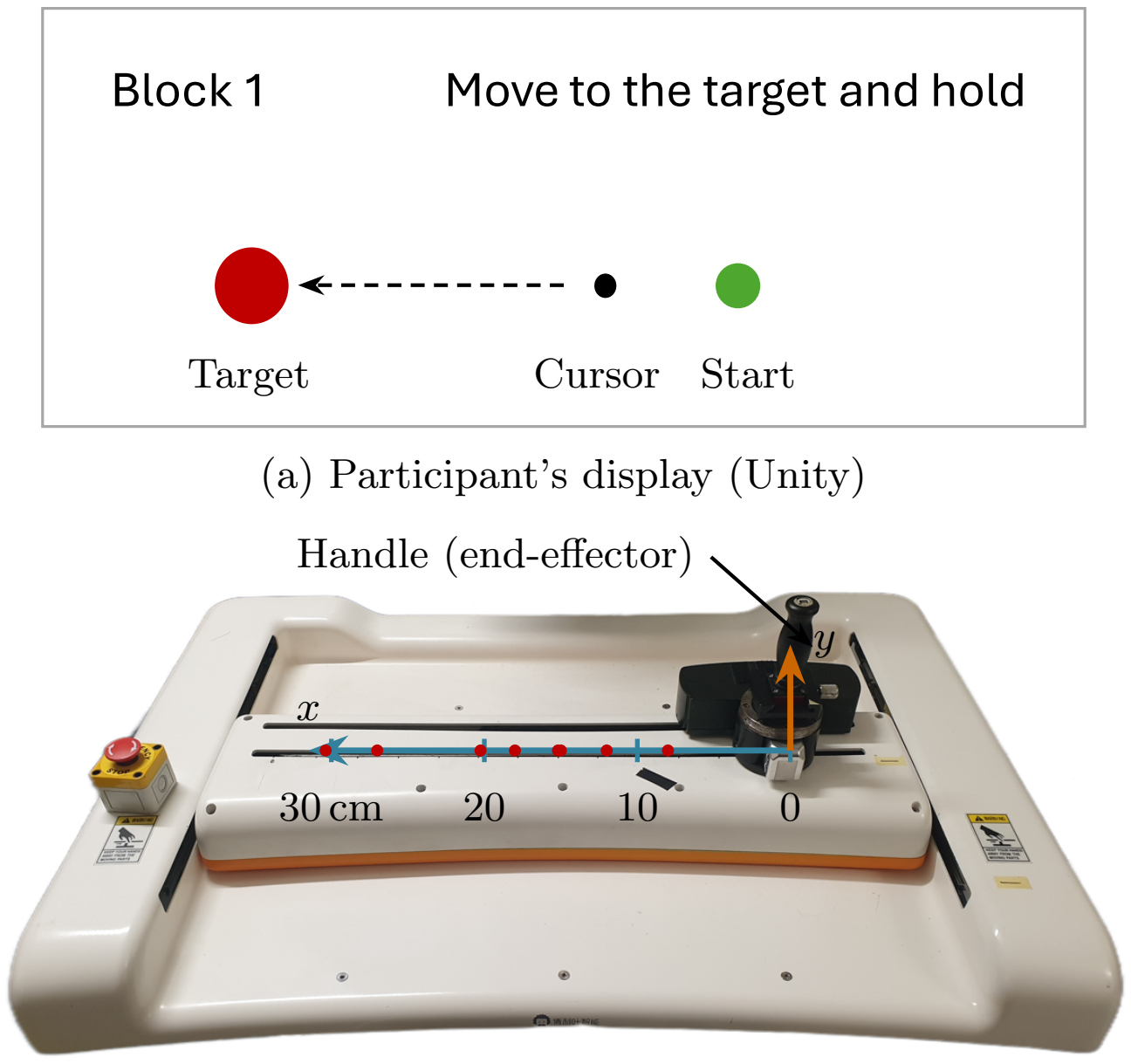

Reaching was conducted using the ArmMotus~M2 (Fourier Intelligence, Shanghai, China, Fig.~\ref{fig:m2setup}), a two-degree-of-freedom planar haptic manipulandum, controlled through the open-source CANopen Robot Controller (CORC) \cite{fong2021}. CORC runs the real-time state machine (calibration, transparent hold, and reset-to-centre) at 500~Hz and streams end-effector position, velocity and interaction force at 40~Hz over a TCP link to a Unity virtual environment. 

The participant sat facing a computer monitor on which Unity rendered a cursor at the handle position and the target to acquire, so that the physical reach on the M2 is performed as a visual pointing task; Unity also logs each trial. Assistance is added per~\eqref{eq:force}, with PD gains set to $K_p=1400$~N\,m$^{-1}$ and $K_d=100$~N\,s\,m$^{-1}$.

Every reach was conducted over the same direction of a single horizontal axis, the $x$-axis of the M2 workspace (Fig.~\ref{fig:m2setup}), so that all trials traversed the same path. 

\subsection{Protocol}
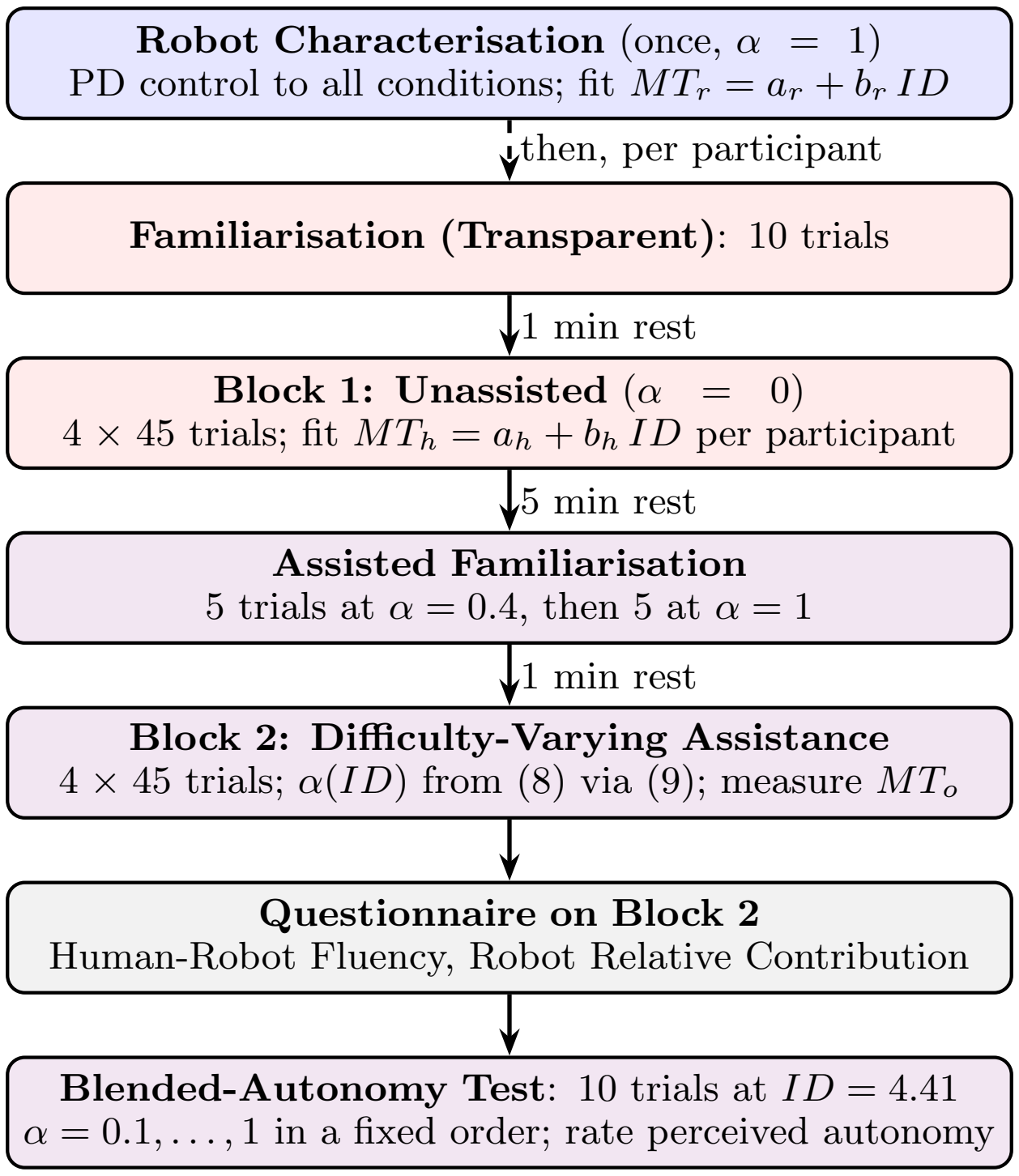
\begin{figure}[t]
\centering
\begin{tikzpicture}[font=\small, node distance=5mm,
  blk/.style={draw, thick, rounded corners, align=center,
    text width=0.94\columnwidth, inner sep=2pt, minimum height=9mm},
  rob/.style={blk, fill=blue!10},
  hum/.style={blk, fill=red!8},
  ast/.style={blk, fill=violet!10},
  qst/.style={blk, fill=black!5},
  fl/.style={-{Stealth[length=2.2mm]}, line width=1pt},
  lab/.style={right, font=\footnotesize, inner sep=2pt}]
  \node[rob] (rc) {\textbf{Robot Characterisation} (once, $\alpha=1$)\\
    PD control to all conditions; fit ${MT_r=a_r+b_r\,ID}$};
  \node[hum, below=of rc] (wu) {\textbf{Familiarisation (Transparent)}: 10 trials};
  \node[hum, below=of wu] (b1) {\textbf{Block 1: Unassisted} ($\alpha=0$)\\
    $4\times45$ trials; fit ${MT_h=a_h+b_h\,ID}$ per participant};
  \node[ast, below=of b1] (aw) {\textbf{Assisted Familiarisation}\\ 5 trials at ${\alpha=0.4}$, then 5 at ${\alpha=1}$};
  \node[ast, below=of aw] (b2) {\textbf{Block 2: Difficulty-Varying Assistance}\\
    $4\times45$ trials; $\alpha(ID)$ from \eqref{eq:alpha_id} via \eqref{eq:force}; measure $MT_o$};
  \node[qst, below=of b2] (q) {\textbf{Questionnaire on Block 2}\\ Human-Robot Fluency, Robot Relative Contribution};
  \node[ast, below=of q] (ba) {\textbf{Blended-Autonomy Test}: 10 trials at ${ID=4.41}$\\
    ${\alpha=0.1,\dots,1}$ in a fixed order; rate perceived autonomy};
  \draw[fl, dashed] (rc) -- node[lab]{then, per participant} (wu);
  \draw[fl] (wu) -- node[lab]{1 min rest} (b1);
  \draw[fl] (b1) -- node[lab]{5 min rest} (aw);
  \draw[fl] (aw) -- node[lab]{1 min rest} (b2);
  \draw[fl] (b2) -- (q);
  \draw[fl] (q) -- (ba);
\end{tikzpicture}
\caption{Experiment protocol. The robot was characterised once on its own and shared across participants, giving the robot Fitts line $MT_r$.
Each participant then completed an initial familiarisation before performing the unassisted Block~1 (giving
$MT_h$). They were then familiarised with the assistance and performed the $\alpha$-varying Block~2. Each block ran as four rounds of 45 trials with one-minute breaks between rounds. A questionnaire
then rated Block~2, before the session ended with a blended-autonomy test in which
$\alpha$ varied and the participant rated their perceived autonomy.}
\label{fig:protocol}
\end{figure}

Fig.~\ref{fig:protocol} depicts the experiment protocol. The robot was first characterised once for all participants: the M2 drove itself to every condition under the regulator~\eqref{eq:pd}, with a single operator's hand resting passively on the handle so that a representative limb inertia and passive impedance were present to fit $MT_r=a_r+b_r\,ID$. Using one characterisation assumes the between-participant variation in passive limb load is small relative to its effect on the $MT_r$ line. 

Each participant then followed Fig.~\ref{fig:protocol}. After ten familiarisation trials on the transparent device, Block~1 measured the unassisted human ($\alpha=0$), and was used to fit $MT_h=a_h+b_h\,ID$ per participant. After an assisted familiarisation of five trials at $\alpha=0.4$ and five at $\alpha=1$, Block~2 then applied difficulty-varying assistance toward a constant target set at the participant's fastest unassisted reach over the tested range so that $MT_o^\star=MT_h(ID_{\min})$. The target gives $\alpha(ID_{\min})=0$, so the easiest condition is reached unassisted, and the weight $\alpha(ID)$ of~\eqref{eq:alpha_id}, clamped to $[0,1]$, rises with difficulty and is applied at each condition through~\eqref{eq:force}. Setting the target per participant gives each the same difficulty reduction relative to their own reaching. 

After Block~2, participants rated the block on 7-point Likert scales using \textcolor{black}{ the subjective fluency scale (Human-Robot Fluency and Robot Relative Contribution) developed and validated by} \cite{hoffman2019fluency} (Table~\ref{tab:questionnaire}). The session ended with a blended-autonomy test of ten trials at condition~12 ($A=20.25$~cm, $W=1$~cm, $ID=4.41$~bits), applying each $\alpha\in\{0.1,0.2,\dots,1\}$ once in a random order (fixed across participants) beginning at $\alpha=1$; 
after each trial the participant rated the autonomy they perceived on a scale from 0 to 1. 

Fifteen $(A,W)$ conditions were chosen so that there were five distinct IDs, spanning $2.81$ to $4.97$~bits, each realised by three different amplitude--width pairs (Table~\ref{tab:conditions}). Holding $ID$ fixed while varying $A$ and $W$ separates the effect of difficulty from that of amplitude or width alone; amplitudes range from $8$ to $30.4$~cm and widths from $0.5$ to $2.5$~cm, within the reachable $x$-workspace. Each block presented the conditions in a randomised order over four rounds of 45 trials, three repetitions of each condition per round, giving 12 repetitions per condition and 180 trials per block.

\begin{table}[t]
\centering
\caption{Reaching conditions. Fifteen amplitude--width pairs realise five
indices of difficulty \eqref{eq:shannon}, three conditions at each $ID$, spanning
$2.81$ to $4.97$~bits. Amplitude $A$ is measured from the
common $0$~cm start point}
\label{tab:conditions}
\begin{tabular}{cccc}
\toprule
Condition & $A$ (cm) & $W$ (cm) & $ID$ (bits) \\
\midrule
$1$  & $12.00$   & $2.000$ & $2.81$ \\
$2$  & $8.00$    & $1.333$ & $2.81$ \\
$3$  & $15.00$   & $2.500$ & $2.81$ \\
\midrule
$4$  & $18.00$   & $2.000$ & $3.32$ \\
$5$  & $12.00$   & $1.333$ & $3.32$ \\
$6$  & $8.00$    & $0.889$ & $3.32$ \\
\midrule
$7$  & $27.00$   & $2.000$ & $3.86$ \\
$8$  & $18.00$   & $1.333$ & $3.86$ \\
$9$  & $12.00$   & $0.889$ & $3.86$ \\
\midrule
$10$ & $27.00$   & $1.333$ & $4.41$ \\
$11$ & $18.00$   & $0.889$ & $4.41$ \\
$12$ & $20.25$   & $1.000$ & $4.41$ \\
\midrule
$13$ & $27.00$   & $0.889$ & $4.97$ \\
$14$ & $30.375$  & $1.000$ & $4.97$ \\
$15$ & $15.1875$ & $0.500$ & $4.97$ \\
\bottomrule
\end{tabular}
\end{table}

\subsection{Measures and Analysis}
Each trial began when the target appeared and ended once the cursor had remained within the target for $1$~s. Movement time was computed as the interval from target onset to first target entry. For each participant and $ID$, $MT$ was summarised as the mean of its 36 trials and normalised by that participant's target $MT_o^\star$, and the robot line was normalised by the same value. 

Fitts-type dependence (H1A) was quantified by a least-squares slope of normalised $MT$ on $ID$ for each participant and block, and the 20 slopes were tested against zero with one-sample $t$-tests. Practice was assessed at $ID_{\min}$, where $\alpha=0$ in both blocks, with a paired $t$-test on the log ratio of Block~2 to Block~1 $MT$. The departure from the target (H1B) was analysed with a linear mixed model of $\ln(MT_o/MT_o^\star)$ with a random intercept for each participant, fitted by REML with Satterthwaite degrees of freedom; the log ratio expresses each departure as a proportion of the participant's own target, since targets differed about two-fold between participants, and effects are reported as $100(e^{b}-1)$\%. The same model, refitted with $ID$ as a fixed effect over the four robot-assisted $ID$s, tested for a residual trend, and the departure was also evaluated against a practice-adjusted target that scales $MT_h$ by the Block~2 to Block~1 ratio at $ID_{\min}$. Finally, shared-control and robot $MT$ were compared in each participant-$ID$ condition to test the convex bound $MT_o\ge MT_r$ of the blend.

The questionnaire (H2) and perceived autonomy (H3) were analysed as described in Sections~\ref{sec:questionnaire} and~\ref{sec:perceived}. Normality was checked with Shapiro--Wilk tests, and tests were two-sided at a significance level of 0.05 unless stated otherwise.

\begin{figure}[!t]
    \centering
    \includegraphics[width=\columnwidth]{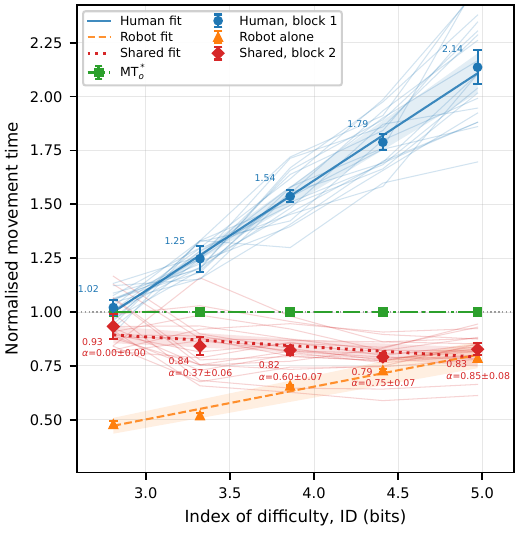}
    \caption{Normalised $MT$ against $ID$ for the unassisted human (Block~1, blue circles), robot alone (orange triangles) and shared control (Block~2, red diamonds). 
    Points are means at each $ID$ with 95\% within-subject confidence intervals; lines show the linear regression of all participants, where shaded bands are the 95\% confidence intervals. The dash-dotted green line is the target $MT_o^\star$ always at 1.0. Labels give the mean shared-control value and the applied autonomy $\alpha$ (mean $\pm$ SD) at each $ID$. Unassisted slope $b=0.512$ per bit; shared-control slope $b=-0.048$ per bit (95\% CI $-0.072$ to $-0.023$).}
    \label{fig:fittsplots}
\end{figure}

\begin{figure}[!t]
\centering
\includegraphics[width=\columnwidth]{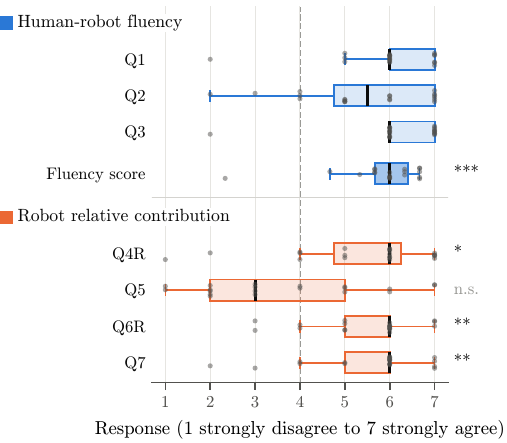}
\caption{Questionnaire responses on the 1--7 scale of \protect\cite{hoffman2019fluency} (item wording in Table~\ref{tab:questionnaire}). Boxes show the median and interquartile range, whiskers extend to 1.5 times the interquartile range, and points are individual responses. Reverse-scored items (R) are inverted, so higher values always mean agreement with the construct, and the fluency score is the mean of Q1--Q3. Markers give Holm-corrected one-sample Wilcoxon signed-rank tests against neutral (4, dashed line) for the fluency score and Q4R--Q7: *~$p<0.05$, **~$p<0.01$, ***~$p<0.001$, n.s.\ not significant.}
\label{fig:questionaire}
\end{figure}

\section{Experimental Results}
\label{sec:results}
All 20 participants completed all blocks, the questionnaire and the blended-autonomy test. 

\subsection{Assisted Performance}
Fig.~\ref{fig:fittsplots} shows the movement time for all participants, normalised by each participant's target $MT_o^\star$ (1.0). Without assistance, $MT$ followed Fitts' law, with normalised fit $-0.437+0.512\,ID$ (mean per-participant $R^2=0.97$): it rose from 1.021 at 2.81~bits (not different from $MT_o^\star$, $p=0.19$) to 2.136 at 4.97~bits, a mean per-participant slope of $+0.512$ per bit (95\% CI 0.469 to 0.554, $t(19)=25.26$, $p<0.001$). The robot alone followed $0.045+0.152\,ID$ ($R^2=0.97$ on the $ID$ means), a slope less than a third of the human's, and was faster than the unassisted participants at every $ID$ (0.48 to $0.79\,MT_o^\star$).

Under shared control the Fitts-type increase was removed: the mean per-participant slope was $-0.048$ per bit (95\% CI $-0.072$ to $-0.023$, $t(19)=-4.04$, $p<0.001$), so movement time no longer rose with difficulty, but was not strictly independent of it. The small negative slope was driven by the easiest $ID$, where $\alpha=0$ and Block~2 $MT$ was already 8.9\% lower than Block~1 $MT$ (paired $t(19)=-4.27$, $p<0.001$), which we attribute to practice between the blocks. H1A is therefore supported in that assistance removed the increase of movement time with difficulty.

The departure from the target was estimated by a linear mixed model of $\ln(MT_o/MT_o^\star)$ with a random intercept for each participant. Shared-control $MT$ was on average 16.3\% lower than $MT_o^\star$ (95\% CI $-19.7$\% to $-12.9$\%, $t(19)=-9.17$, $p<0.001$), supporting H1B. Across the four robot-assisted $ID$s alone the shortfall was 18.5\% (95\% CI $-22.5$\% to $-14.3$\%, $t(19)=-8.53$, $p<0.001$), with no trend across them ($-1.2$\% per bit, 95\% CI $-3.5$\% to $1.0$\%, $p=0.278$). The shortfall remained against a practice-adjusted target, $\alpha MT_r+(1-\alpha)\,0.911\,MT_h$, at every assisted $ID$ ($-8.3$\%, $p=0.014$, at 3.32~bits and $-14.5$\% to $-18.1$\%, $p<0.001$, at the three hardest). In 14 of the 100 participant-$ID$ condition mean values, from 8 participants, shared control was faster than the robot alone (one at 3.86~bits, six at 4.41~bits and seven at 4.97~bits), although on average it remained slower than the robot ($+9.7$\% at 4.41~bits, $p=0.006$; $+6.2$\% at 4.97~bits, $p=0.082$).

\begin{figure}[!t]
\centering
\includegraphics[width=\columnwidth]{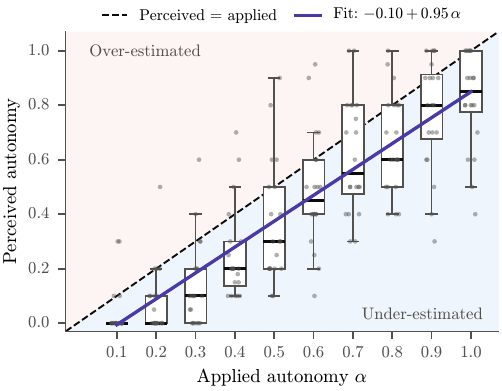}
\caption{Perceived autonomy $\hat{\alpha}$ against applied autonomy $\alpha$ in the blended-autonomy test ($n=20$, one trial per level at $ID=4.41$~bits). Boxes show the median and interquartile range, whiskers extend to 1.5 times the interquartile range, and points are individual ratings. Ratings above the dashed identity line over-estimate the applied autonomy and those below under-estimate it; the solid line is the mixed-effects calibration fit.}
\label{fig:perceived}
\end{figure}

\subsection{Questionnaire}
\label{sec:questionnaire}

\begin{table*}[t]
\centering
\caption{Questionnaire results for the items of \protect\cite{hoffman2019fluency}. Responses range from 1 (strongly disagree) to 7 (strongly agree), with 4 neutral; reverse-scored items (R) are inverted, so agreement means disagreement with the statement as worded. The fluency score is the mean of Q1--Q3; the robot relative contribution items are reported separately, as that scale was not internally consistent. $p_{\mathrm{Holm}}$: two-tailed one-sample Wilcoxon signed-rank test against neutral, Holm-corrected across the fluency score and the four contribution items; $r_{\mathrm{rb}}$: rank-biserial correlation.}
\label{tab:questionnaire}
\small
\setlength{\tabcolsep}{6pt}
\renewcommand{\arraystretch}{1.15}
\begin{tabular}{@{}l>{\raggedright\arraybackslash}p{0.55\textwidth}rcrr@{}}
\toprule
Item & Statement & Median & IQR & $p_{\mathrm{Holm}}$ & $r_{\mathrm{rb}}$ \\
\midrule
\multicolumn{6}{@{}l}{\textit{Human-robot fluency} (Q1--Q3, $\alpha_{\mathrm{C}}=0.69$)} \\
Score & Fluency: mean of Q1--Q3 & 6 & [5.67, 6.42] & $<0.001$ & 0.95 \\
Q1 & The human-robot team worked fluently together. & 6 & [6, 7] &  &  \\
Q2 & The human-robot team's fluency improved over time. & 5.5 & [4.75, 7] &  &  \\
Q3 & The robot contributed to the fluency of the interaction. & 6 & [6, 7] &  &  \\
\midrule
\multicolumn{6}{@{}l}{\textit{Robot relative contribution} (Q4--Q7, $\alpha_{\mathrm{C}}=-0.57$, items analysed separately)} \\
Q4R & I had to carry the weight to make the human-robot team better. & 6 & [4.75, 6.25] & 0.017 & 0.71 \\
Q5 & The robot contributed equally to the team performance. & 3 & [2, 5] & 0.41 & $-0.22$ \\
Q6R & I was the most important team member on the team. & 6 & [5, 6] & 0.002 & 0.92 \\
Q7 & The robot was the most important team member on the team. & 6 & [5, 6] & 0.003 & 0.87 \\
\bottomrule
\end{tabular}
\end{table*}

Internal consistency was assessed with Cronbach's alpha, $\alpha_{\mathrm{C}}$, for each scale of \cite{hoffman2019fluency} (Table~\ref{tab:questionnaire}, Fig.~\ref{fig:questionaire}). The three-item human-robot fluency scale gave $\alpha_{\mathrm{C}}=0.69$, so its items were averaged into a single fluency score. The four-item robot relative contribution scale was not internally consistent ($\alpha_{\mathrm{C}}=-0.57$), largely because Q5 (``The robot contributed equally to the team performance'') split participants, so its items were analysed separately. 

As the responses were ordinal and skewed toward agreement (Shapiro--Wilk $p<0.05$), the fluency score and each contribution item were compared with the neutral midpoint (4) by a two-tailed one-sample Wilcoxon signed-rank test with Holm correction. Participants rated the interaction as fluent (median 6 on the 1 to 7 scale, $p<0.001$, $r_{\mathrm{rb}}=0.95$), supporting H2. They disagreed that they had carried the team (Q4R) or been its most important member (Q6R), and agreed that the robot had been (Q7), all $p\leq0.017$, while Q5 was split (35\% agree, 55\% disagree, $p=0.41$). Participants therefore found the shared control fluent and perceived the robot, rather than themselves, as the dominant contributor.

\subsection{Perceived Autonomy}
\label{sec:perceived}
Fig.~\ref{fig:perceived} compares the autonomy each participant perceived in the blended-autonomy test, $\hat{\alpha}$, with the applied weight $\alpha$. A linear mixed-effects model of $\hat{\alpha}$ on $\alpha$, with a random intercept and slope for each participant, gave $\hat{\alpha}=-0.10+0.95\,\alpha$. The slope (95\% CI 0.81 to 1.09) did not differ from one ($p=0.49$), so a change in autonomy was perceived at its true size; however, the intercept (95\% CI $-0.18$ to $-0.03$, $p=0.008$) placed the whole response below the identity line. Participants therefore tracked the level of autonomy but consistently under-estimated it by about 0.1, so H3 is supported for changes in autonomy but not for its absolute level.

\section{Discussion}
Difficulty-indexed assistance removed the increase of movement time with difficulty seen without assistance (H1A). The linear blend did not, however, predict the level at which movement time settled. Assisted reaches were on average 16.3\% faster than the target $MT_o^\star$, so the passive-human blend~\eqref{eq:blend} is rejected in the direction anticipated by \cite{pan2025fitts} (H1B). On average the assisted times stayed above the robot line. However, in 14 of the 100 participant-$ID$ condition mean values, from 8 participants, shared control was faster than the robot alone, which no convex combination of the two can produce.

The sign of the departure points to its source. The target assumes the participant supplies the complementary share $(1-\alpha)\,u_h$ of the full regulator~\eqref{eq:pd}; supplying more, as feedforward drive, settles the reach sooner ($MT_o<MT_o^\star$), whereas slacking or passive loading by the limb would slow it. The faster reaching we observed is therefore consistent with feedforward engagement, and the perceived autonomy points the same way: the under-estimation of the robot's share could be caused by participants attributing part of its motion to their own effort.

Part of the shortfall could also be caused by practice. At the easiest $ID$, where $\alpha=0$, Block~2 was already 8.9\% faster than Block~1, so the target set from Block~1 was slower than a practised participant's own reach. However, the same improvement could also reflect participants being drawn to move faster by the surrounding assisted trials. Either effect favours the feedforward conclusion. Practice alone, however, acts only on the human share $(1-\alpha)\,MT_h$ of the blend, so its contribution should shrink as $\alpha$ rises; yet against a practice-adjusted target the shortfall remained, at 8.3\% at 3.32~bits and 14.5 to 18.1\% at the three hardest $ID$s. This pattern favours an active departure that grows with assistance over practice alone, although the present design cannot fully separate the two.

Participants rated the interaction as fluent and the robot as the most important team member (H2), and perceived changes in $\alpha$ at their true size while under-estimating its level (partly supporting H3). The two are consistent: judging the robot the most important member is a relative comparison, whereas $\hat{\alpha}$ is an absolute estimate of its share. Difficulty-indexed assistance was therefore legible to its users, even if they credited themselves with more of the motion than they supplied.

The departure has a direct consequence for the rehabilitation setting that motivates the work. Weights computed from the passive blend over-assist an engaged user, delivering a task easier than intended, here by up to about 16\%, and pulling them below the challenge their skill warrants \cite{nakamura2002flow}. Holding a user within the flow channel therefore requires the operator's own contribution to be folded back into the arbitration. 

This study has \textcolor{black}{six} main limitations. (i) Block~2 always followed Block~1, so practice, or a context effect of the assisted trials, is confounded with the assistance; a counterbalanced order or a final unassisted block would separate them. (ii) $\alpha(ID)$ is a function of $ID$ by construction, so the effects of $\alpha$ and $ID$ on movement time cannot be separated within Block~2. (iii) Participants were 20 healthy adults aged 20 to 30; transfer to patients and to the retention of skill is untested. \textcolor{black}{(iv) The experiment was conducted on one robot, where the weights are likely device- and gain-specific.} (v) Without electromyography or a measure of human effort, feedforward drive cannot be separated from co-contraction, the robot line was characterised once with one operator's passive limb, and the robot relative contribution scale was not internally consistent, so only its individual items could be interpreted. \textcolor{black}{(vi) A notable design constraint is the single axis reaching motion. Here we note that although the ballistic movement is naturally in a single direction, that direction is potentially different for each reach, where each direction could be modelled for.} 

These limitations point to the future steps. As the operator's contribution is not the passive share the blend assumes, the most direct extension is an adaptive arbitration that accounts for the human's current action, for example by updating $\alpha$ online from the realised MT or the measured interaction force, so that the target time is met. Recording electromyography and perceived difficulty would decompose the departure into its active and passive parts, and a multi-session study with washout and retention would test whether holding a learner in the flow channel improves skill, rather than only aiding the reach in the moment. \textcolor{black}{Different reaching axes and the effect of other devices could also be investigated.}

\section{Conclusion}
This study indexed haptic assistance to task difficulty rather than tracking error, placing the human and the assisting robot on a common Fitts' law scale for reaching tasks. Assuming a linear blend of their movement times gave a closed-form arbitration weight, and choosing a target that makes movement time invariant to difficulty yielded an assist-as-needed controller whose passive-human assumption is itself falsifiable. In a study with 20 participants on a planar manipulandum, the assistance removed the Fitts increase of movement time with difficulty as designed. However, movement time settled about 16\% below the target, and in some conditions below the robot alone, consistent with participants adding feedforward drive rather than slacking, with practice between blocks accounting for only part of this departure. Participants found the interaction fluent, judged the robot as the main contributor, and tracked the applied autonomy consistently while under-estimating its level. For rehabilitation, these results indicate that weights computed from a passive blend would over-assist an engaged patient and pull them below the flow channel the assistance is meant to hold. Future work will therefore consider adaptive arbitration that accounts for the human's current contribution, together with measures of perceived difficulty and effort, so that the target difficulty is met rather than overshot.

\section*{Acknowledgements}
This work is supported by the Australian Research Council Projects DP260101082 and FT250100459. We thank the participants who took part in this study, the University of Melbourne Human Robotics Lab and the developers of CORC. Generative AI (Claude, Anthropic) assisted in drafting and editing the text, the statistical analysis code and figure preparation; the authors verified all content and take full responsibility for it.

\bibliographystyle{named}
\bibliography{references}

\end{document}